# Adapting Word Embeddings to New Languages with Morphological and Phonological Subword Representations

**Aditi Chaudhary, Chunting Zhou, Lori Levin,
Graham Neubig, David R. Mortensen, Jaime G. Carbonell**
Language Technologies Institute, Carnegie Mellon University
{aschaudh,ctzhou,lsl,gneubig,dmortens,jgc}@cs.cmu.edu

## Abstract

Much work in Natural Language Processing (NLP) has been for resource-rich languages, making generalization to new, less-resourced languages challenging. We present two approaches for improving generalization to low-resourced languages by adapting continuous word representations using linguistically motivated subword units: phonemes, morphemes and graphemes. Our method requires neither parallel corpora nor bilingual dictionaries and provides a significant gain in performance over previous methods relying on these resources. We demonstrate the effectiveness of our approaches on Named Entity Recognition for four languages, namely Uyghur, Turkish, Bengali and Hindi, of which Uyghur and Bengali are low resource languages, and also perform experiments on Machine Translation. Exploiting subwords with transfer learning gives us a boost of +15.2 NER F1 for Uyghur and +9.7 F1 for Bengali. We also show improvements in the monolingual setting where we achieve (avg.) +3 F1 and (avg.) +1.35 BLEU.

## 1 Introduction

Continuous word representations have demonstrated utility in state-of-the-art neural models for several NLP tasks, such as named entity recognition (NER; Ma and Hovy (2016)), machine reading (Tan et al., 2017), sentiment analysis (Tang et al., 2016; Yu et al., 2018), and machine translation (MT; Qi et al. (2018)). While the training of these word vectors does not rely on explicit human supervision, their quality is highly contingent on the size and quality of the unlabeled corpora available. There are over 7000 languages in the world (Hammarström et al., 2018), and corpora with sufficient size and coverage are available for just a handful, making it unclear how these methods will perform in the more common low-resource setting.

Disheartening though this high dependence on resources sounds, several efforts (Adams et al., 2017; Haghighi et al., 2008; Bharadwaj et al., 2016; Mayhew et al., 2017) have shown considerable performance gains across different tasks in the low resource setting by transferring knowledge from related high-resource languages. Most existing approaches for learning cross-lingual word embeddings (Ruder, 2017) either extend the monolingual objective function by adding a cross-lingual regularization objective which is then jointly optimized or use mapping-based approaches to align similar words across languages. These post-hoc coordination methods rely on bilingual lexicons or parallel corpora, which are typically of limited quantity and uncertain quality.

In this paper, we take a different task: focusing instead on the *similarity of the surface forms, phonology, or morphology* of the two transfer languages. Specifically, inspired by Ling et al. (2015), who demonstrate the effectiveness of character-level modeling for knowledge sharing in multilingual scenarios, we propose two approaches to transfer word embeddings using different types of linguistically-inspired subword-level information. Both approaches focus on mapping the low resource language embeddings closer to those of the high resource language and are executed using two different training regimes. We explore the effect of different subword units— characters, lemmas, inflectional properties, and phonemes— as each one offers a unique linguistic insight, discussed more in Section 3. Our proposed approaches do require language specific resources, but importantly do not depend on cross-lingual resources and achieve considerable performance gains over existing methods which do.

We evaluate our proposed approach on two downstream tasks: NER, which deals with detecting and classifying Named Entities (NEs) into pre-defined categories (Nadeau and Sekine, 2007), and MT to English. For the purposes of error analysis and discussion, we focus on the NER task in particular. NEs are typically noun phrases and occur rarely in the corpus, making the generalization across types and domains difficult. We chose NER as our test bed because word vectors have a direct impact on NER model performance— as suggested by (Ruder, 2017) and observed by us in Table 3, where the model without any pre-trained embeddings scores an average of 18 F1 points less. It thus provides a transparent way to measure the effectiveness of different subword units.

This paper makes the following contributions:

1. We show that embeddings trained on subword representations yield better task performance than those trained only on whole words. This is especially true in a transfer setting, where subword representations also outperform a word alignment based method. We further show that embeddings trained on morphological representations often outperform those trained only on whole words.

2. We demonstrate that training embeddings on character-based phonemic representations presents substantial performance advantages over training on orthographic characters in some transfer settings, e.g. when there are script differences across languages. These advantages are in addition to those from morphological representations (lemmas and morphological properties).

3. We produce continuous representations for each subword unit, giving researchers the ability to use them in their own tasks as they see fit. The code [1] for training word embeddings and the embeddings [2] which produced the best results are publicly available. We also release morphological analyzers for Hindi and Bengali[3].

## 2 Skipgram Objective

The two most popular training objectives for monolingual word embeddings are the skip-gram and continuous-bag-of-words (CBOW), introduced by Mikolov et al. (2013a). The skipgram model attempts to predict the context surrounding a word, given the word itself whereas CBOW predicts the word given its context. Formally, given a corpus having a sequence of words $w_1, w_2, \cdots, w_T$, the skip-gram model maximizes the following log-likelihood:

$$\sum_{i=1}^{T} \sum_{v \in C_i} \log p(v|w_i) \quad (1)$$

where $C_i$ are the context tokens, within a specified window of the focus word $w_i$ and $p(v|w_i)$ is the probability of observing context word $v$ given focus word $w_i$. The skipgram was originally defined using the softmax function:

$$p(v|w_i) = \frac{e^{s(v,w_i)}}{\sum_{j=1}^{W} e^{s(w_i,j)}} \quad (2)$$

where $s$ is a scoring function mapping $v$ and $w_i$ to $\mathbb{R}$. The summation in the denominator is over the entire vocabulary $W$ which makes this formulation computationally inefficient as cost of gradient computation is proportional to $W$ which is quite large ($\sim 10^6$). Mikolov et al. (2013b) hence employ negative sampling to make this computation efficient and robust (Levy et al., 2015) and give better representations for infrequent words[4], which is crucial for the low resource settings. Negative sampling represents the above objective function (Equation 1) using a binary logistic loss as shown below:

$$\sum_{i=1}^{T} \left( \sum_{w_c \in C_i} l(s(w_i, w_c)) + \sum_{w_n \in N_i} l(-s(w_i, w_n)) \right) \quad (3)$$

where $N_i$ are the negative words sampled randomly from vocabulary and $l$ is the log-sigmoid function. The scoring function $s$ is a dot product similarity function given by $s(w_i, w_c) = \mathbf{u}_{w_i}^\top \mathbf{v}_{w_c}$ where $\mathbf{u}_{w_i}$ and $\mathbf{v}_{w_c}$ are the embeddings of the focus word and its context word respectively.

---
[1] https://github.com/Aditi138/Embeddings
[2] https://github.com/Aditi138/Embeddings/tree/master/embeddings_released
[3] https://github.com/dmort27/mstem

## 3 Subword Representation

Mikolov et al. (2013b)'s model fails to capture internal structure of words and does not generalize for out of vocabulary words that may share morphemes with in-vocabulary words. The problems of this method are particularly salient for

---
[4] https://code.google.com/archive/p/word2vec/

| | |
|---:|:---|
| graphemes | ⟨ﻗﺎرﯨﻴﺎﻟﻤﺎﻳﺪۇ⟩ |
| phonemes | /qarijalmajdu/ |
| morphemes | /qari-jal-ma-jdu/ |
| lemma+tag | qari+Verb+Pot+Neg+Pres+A3sg |
| gloss | 's/he can't care for' |

Figure 1: Representations of a word in Uyghur

| | |
|---:|:---|
| phoneme ngrams | $x_{<qa} + x_{qar} + ... + x_{majdu>}$ |
| lemma | $x_{\text{qari}}$ |
| morphemes | $x_{Verb} + x_{Pot} + ... + x_{A3sg}$ |

Figure 2: Vector representations of a word in Uyghur

morphologically rich languages such as Turkish, Uyghur, Hindi, and Bengali. Although, given a large enough training corpus, most or all morphological forms of a lexeme (of which there may be many) could theoretically learn to have similar vector representations, it will be vastly more data efficient if we can take into account regularities of their form to model morphology explicitly. We explore the following methods for doing so:

**Orthographic units:** Wieting et al. (2016) and Bojanowski et al. (2016) show the utility of character-level modeling by representing the focus word $w_i$ as a set of its character ngrams, denoted by $\mathbf{u}_{w_i} = \frac{1}{|G|} \sum_{g \in G} \mathbf{x}_g$, where $G$ is the set of character ngrams and $\mathbf{x}_g$ is the vector representation of ngram $g$. Such representations capture morphological information in a brute-force but principled fashion—words that share the same morpheme are more likely to share the same character ngrams than words that do not.

**Morphological units:** Previous work has found that morphological relationships between words can be captured more directly if embeddings are trained on morphological representations (Luong et al., 2013; Botha and Blunsom, 2014; Cotterell and Schütze, 2015). Avraham and Goldberg (2017) explicitly model lemmas (stems or citation forms) and morphological properties (the sets of which are sometimes called "tags") for training the word embeddings. Lemmas capture information about the lexical identity of a word and are closely correlated with the semantics of a word; tags capture information about the syntactic context of a word. See Figure 1 for an example. We take inspiration from the above work in adapting these subword units for cross-lingual transfer.

**Phonological units:** Subword units other than tags might seem to be of no use in closely-related languages with different scripts (such as Serbian and Croatian). Following Bharadwaj et al. (2016), we convert text from its orthographic form into a phonemic representation, stated in terms of the International Phonetic Alphabet (IPA). We then train embeddings on this representation. This means that, roughly speaking, morphemes that sound the same will be represented in the same way across languages.

## 4 Cross-lingual Transfer

In this section we discuss in detail both our approaches for cross-lingual transfer along with the relevant baselines.

### 4.1 Proposed Approach

We propose to use phoneme ngrams, represented using IPA, in addition to the lemma and morphological tags, to enable effective transfer across languages. Tsvetkov and Dyer (2016) demonstrate the effectiveness of projecting words from orthographic space to phonemic space as related languages often share similar phonological patterns. More formally, let $P_w$ be the set of linguistic properties of a word consisting of the phoneme ngrams ($\mathbf{I}_g$), lemma ($\mathbf{L}$) and individual morphological tags ($\mathbf{M}_m$). The focus word is then represented as the average sum of its linguistically motivated subword units:

$$\mathbf{v}_{w_c} = \frac{1}{|P_{w_c}|} \sum_{p \in P_{w_c}} \mathbf{x}_p$$

where $\mathbf{x}_p$ is the vector representation of subword unit $p$ of word $w_c$. The average operation is important to remove any bias towards words having too many or too few subword units. For instance, the Uyghur word in Figure 1 is represented using its phoneme-ngrams ranging from 3-grams to 6-grams, lemma and morphological tags as shown in Figure 2. Avraham and Goldberg (2017) instead encode the different morphological inflections as one tag, so that *Verb+Pot+Neg+Pres+A3sg* would be encoded as $x_{Verb+Pot+Neg+Pres+A3sg}$. We encode each property in a tag separately to avoid data sparsity issues and empirically find this approach to perform better.

We present two training regimes for transferring knowledge from a related language, namely **CT-**

**Joint** and **CT-FineTune** by explicitly incorporating the subword units. We hypothesize that having word representations of both languages lying in a similar space will aid the low resource language in leveraging resources from the high resource language, including annotations for the downstream task. These two regimes are described below:

**CT-Joint:** This model explicitly maps the word representations of the two languages into the same space by training simultaneously on both. This is achieved simply by combining the corpora of both the high-resource and the low-resource language and training jointly using the skip-gram objective, discussed above. The central intuition is as follows: once two related languages are placed in the same phonological and morphological space, they will share many subword units in common and this will make joint training profitable. Duong et al. (2016) and Gouws et al. (2015) have previously shown the advantages of joint training and we observe this to be true in our case as well.

**CT-FineTune:** This model implicitly maps the word representations of the two languages into the same space. The model attempts this by taking the learned continuous representations of the high resource subword units, referred to by $\mathbf{x}_{SWU}^{Hi}$, and uses them to initialize the model for the low resource language. The model is first trained using all subword units on the high resource language and the learned representations are then used for initializing the subword units for the low resource language. To elucidate which pretrained subword helped the most on the low resource language, we use the same model for different experiments, which is trained using all subword units—phoneme-ngrams, lemma and morphological properties. The linguistic intuition behind CT-FineTune is similar to that behind CT-Joint. This idea of transferring parameters from high resource language has been previously explored by Zoph et al. (2016) for low resource neural machine translation which showed considerable improvement.

## 5 Evaluation

In this section, we first describe the model setup for training word embeddings followed by details on NER and MT experiments.

### 5.1 Implementation details

We base our model on the C++ implementation of *fasttext*[5] (Bojanowski et al., 2016) with modifications as described above.

**Data:** We represent a word in the training corpus using the format presented by Avraham and Goldberg (2017). For instance, the Uyghur word in Figure 1 is represented as follows: phoneme **ipa:** *qarijalmajdu*, lemma **l:***qari*, and morphological inflections **m:***Verb+Pot+Neg+Pres+A3sg*. We consider phoneme-ngrams ranging from 3-grams to 6-grams and append a special start symbol < and end symbol > to the word. We discard unigrams and bigram ngrams on the assumption that they don't contribute much to the word.

**Linguistic properties:** We experiment with different subword units for both the transfer setting and the monolingual setting. We use the orthography-to-IPA tool *Epitran* (Mortensen et al., 2018) to obtain the phonemic representations. The lemmas and morphological properties for a word in context are obtained using a rule-based morphological analyzer in such a fashion as to produce tags similar to the high resource language. For Turkish we use the morphological disambiguator developed by (Shen et al., 2016), which in turn is based on an FST-based morphological analyzer developed by Oflazer (1994). For Uyghur, we took a (parser combinator based) morphological analyzer that had been developed for a DARPA LORELEI evaluation and modified it to output part-of-speech tags and to use a property set that was as close as possible to that of the Oflazer Turkish analyzer. The analyzer for Turkish produces 116 inflectional properties and for Uyghur we get 54 properties, of which 64% are shared with Turkish. Unfortunately, we did not have access to existing morphological analyzers for Hindi or Bengali. Many Hindi morphological analyzers exist, but they are not typically released publicly (Malladi and Mannem, 2013; Goyal and Lehal, 2008). We developed our own analyzers using a stemmer-like framework[6] over a span of few weeks (2-3), which gave 8 unique morphological tags for Hindi and 10 for Bengali (for both languages, noun inflection only) of which just 2 were shared with Hindi.

Morphologically speaking, we only use inflec-tional properties. For most languages, we considered derivational affixes to be part of the stem, since they change the meaning and grammatical category of the word rather than simply expressing syntactic information. An exception to this was Turkish, where the available morphological analyzer segments all affixes off from the root. However, even there we confined our use of morphological properties to inflectional properties. Derivational affixes display scopal behavior; since we wanted to treat the morphological properties of a word as a set, rather than a sequence, we were required to choose this option.

---
[5] https://github.com/facebookresearch/fastText/
[6] https://github.com/dmort27/mstem

**Hyperparameters:** During training, we consider context tokens within a window size 3 of the focus word and we sample 5 negative examples from the vocabulary. We chose a window size of only 3 based on the fact that we are working with morphologically rich languages with a relatively high information to token ratio (otherwise a window size of 5 may be more appropriate). Subword units are initialized with uniform samples from $\left[\frac{-1}{dim}, \frac{1}{dim}\right]$ where $dim = 100$. We use the same training regime as Bojanowski et al. (2016). For CT-FineTune, instead of uniform samples we initialize the subword units of the low resource language from the learnt $\mathbf{x}_{SWU}^{Hi}$.

| Lang. | Train | Dev | Test |
|---|---|---|---|
| Turkish | 3376 | 1126 | 1126 |
| Uyghur | 1822 | 240 | 2448* |
| Hindi | 3974 | 497 | 497 |
| Bengali | 1908 | 53 | 7012 |

Table 1: Sentences in train/dev/test set for NER. (*Unsequestered set. The full test set has 12,546 sentences.)

### 5.2 Baselines

For comparison, we train multilingual embeddings using MultiCCA (Ammar et al., 2016) as our baseline. It employs canonical correlation analysis by projecting multiple languages in the same shared space of one language, also referred to as multi-language space. This method learns linear projections for each language into this common language space using bilingual lexicons. English is used as a common vector space due to availability of corresponding bilingual lexicons between English and each of our languages. For a fair comparison, we run MultiCCA on monolingual embeddings trained with different subword units. We use 100 dimension (Bojanowski et al., 2016) embeddings for English.

For NER, we also compare with Bharadwaj et al. (2016) who use a neural attention model over phonological features and report the best performance for Turkish using transfer from Uzbek and Uyghur, and Mayhew et al. (2017) who use a cheap translation method to translate training data from high-resource language into the low-resource language and report best NER results for Uyghur, as part of the LORELEI program. Our work differs from these primarily on two fronts: a) it is independent of the downstream task and can easily be adapted across various tasks, and b) it doesn't require parallel corpora or bilingual dictionaries. For our monolingual experiments, we compare our proposed approach with models using subword representations—Bojanowski et al. (2016) and Avraham and Goldberg (2017).

### 5.3 Named Entity Recognition Task

We use state-of-the-art NER architecture (Ma and Hovy, 2016) as our model for evaluation. The task is to identify NEs and categorize them into four types. Since this is a supervised model, the performance is highly contingent on the quality of labeled data. F1 scores are used as the evaluation metric.

#### 5.3.1 Experiments

We conduct the two main sets of NER experiments,

1. Transfer experiments on the low resource languages—Uyghur and Bengali—using Turkish and Hindi as the high resource languages respectively. We show results using both our proposed models, CT-Joint and CT-FineTune.

2. Monolingual experiments on all four languages: Uyghur, Turkish, Bengali and Hindi. We do an ablation study using different combinations of subword units.

These language pairs were chosen partly out of convenience—the data were available to us as part of the DARPA LORELEI program—and partly because they satisfied certain deeper desiderata. Turkish and Uyghur are fairly closely related to one another, as are Hindi and Bengali. Despite this relationship, the members of both pairs are written in different scripts (Roman and Perso-Arabic; Devanagari and Bengali). Finally, all four languages are morphologically rich, especially Turkish and Uyghur. These qualities allow us to showcase the value of embeddings with subword units.

| Model | subword units | Uyghur | Bengali |
|---|---|---|---|
| CT-Joint | phoneme-ngrams + lemma + morph | 55.00 | **60.33** |
| | phoneme-ngrams + lemma | **56.20** | 59.63 |
| | phoneme-ngrams | 54.90 | 58.50 |
| | phoneme | 51.30 | 53.75 |
| | char-ngrams + lemma + morph | 50.20 | 55.10 |
| | char-ngrams + lemma | 48.20 | 53.83 |
| | char-ngrams | 49.60 | 52.77 |
| | word | 51.80 | 53.69 |
| CT-FineTune | phoneme-ngrams + lemma + morph | 48.60 | 56.19 |
| | lemma + morph | 52.80 | 57.72 |
| | phoneme-ngrams + lemma | 51.00 | 56.83 |
| | phoneme-ngrams | 50.50 | 57.69 |
| | phoneme | 49.20 | 59.86 |
| MultiCCA (Baseline) | char-ngrams + lemma + morph | 41.00 | 50.63 |
| | char-ngrams + lemma | 43.10 | 50.63 |
| | char-ngrams | 45.80 | 38.06 |
| | word | 42.70 | 45.86 |

Table 2: Transfer experiments on NER. Metric F1 (out of 100%). Uyghur transfer is from Turkish; Bengali transfer is from Hindi

**Data Preprocessing:** We use data, comprised of unlabeled corpora, English bilingual dictionaries, annotations, from the Linguistic Data Consortium (LDC) language packs—Turkish and Hindi [7], Bengali[8], from which we generate train-dev-test splits. Uyghur data was released as part of LoReHLT16 task, organized by NIST [9] under the aegis of DARPA, and training annotations were acquired using native speakers as part of the task. For Uyghur we evaluate on an unsequestered set consisting of 199 annotated evaluation documents, released by NIST. For Turkish, Hindi and Bengali, we create our own train-dev-test splits (Table 1). The exact documents from which Mayhew et al. (2017) and Bharadwaj et al. (2016) created their test set is not apparent. The Uyghur corpus has 27 million tokens and the Turkish corpus has about 40 million tokens. Although Bengali is widely-spoken and the unlabeled corpus contains more than 140 million tokens, there are very few named entity annotations available, making it a low-resource language for the purposes of this exercise. To have a fair experimental setup across language pairs, we sub-sample the Bengali and Hindi corpora to have comparable corpus sizes with Uyghur and Turkish respectively. We also up-sample the low resource data for both unlabeled corpora and NER annotations, so the model doesn't become biased towards the high resource language.

**NER model setup:** We train the model using 100-dimensional word embeddings, pre-trained using the above discussed strategies, and use hidden dimension of size 100 for each direction of the LSTM. Stochastic gradient descent was used as the optimizer with a learning rate of 0.015. Dropout of 0.5 was used in the LSTM layer to prevent over-fitting. Uyghur and Turkish were trained for 100 epochs, Bengali and Hindi converged after 70 epochs.

### 5.3.2 Results and Discussion

**Transfer Experiments:** From Table 2 we note that our CT-Joint model trained with phoneme-ngrams, lemma, and morphological tags outperforms the MultiCCA baseline by a significant margin. MultiCCA strongly depends on bilingual dictionaries which is possibly why it performs poorly in our low resource setting, where these dictionaries are not of high quality. The advantage of phoneme-ngrams over char-ngrams is quite apparent here. *phoneme-ngrams+lemma+morph* performs +5.2 F1 points better than *char-ngram+lemma+morph* for both Uyghur and Bengali, and similar increase is observed across other combinations, the only exception being the *phoneme* case for Uyghur which performed -0.5 F1 with respect to its counterpart *word*.

---
[7]LDC2014E115, LDC2017E62, http://www.cfilt.iitb.ac.in/iitb_parallel/
[8]LDC2017E60, LDC2015E13
[9]https://www.nist.gov/

| Model | subword units | Turkish | Uyghur | Hindi | Bengali |
|---|---|---|---|---|---|
| Ours | Char-ngrams + Lemma + Morph | 68.06 | **52.50** | 73.15 | **52.77** |
|  | Char-ngrams + Lemma | **68.61** | 52.40 | 73.37 | 52.09 |
|  | Char-ngrams + Morph | 67.97 | 47.80 | **73.46** | 52.06 |
| prop2vec | Word + Lemma | 66.52 | 46.00 | 71.82 | 50.03 |
|  | Word + Morph | 64.45 | 46.00 | 71.52 | 49.27 |
|  | Word + Lemma + Morph | 68.46 | 47.70 | 70.51 | 48.16 |
| fastText | Char-ngrams | 66.81 | 50.80 | 72.67 | 52.10 |
| word2vec | Word | 62.85 | 46.80 | 72.04 | 49.83 |
| Random | No embedding | 58.94 | 31.30 | 59.89 | 21.25 |

Table 3: NER results for monolingual experiments. Metric F1 (out of 100%)

| Model | Uyghur* (unseq.) | Uyghur* | Turkish | Bengali |
|---|---|---|---|---|
| Ours | **56.20** | **56.00** | **68.61** | **60.33** |
| Bharadwaj et al. (2016) | – | 51.2 | 66.47 | – |
| Mayhew et al. (2017) | 51.32 | 55.6 | 53.44 | 45.70 |

Table 4: Comparison with previous work using data released by DARPA LORELEI. Metric F1 (out of 100%) *Official NIST scores.

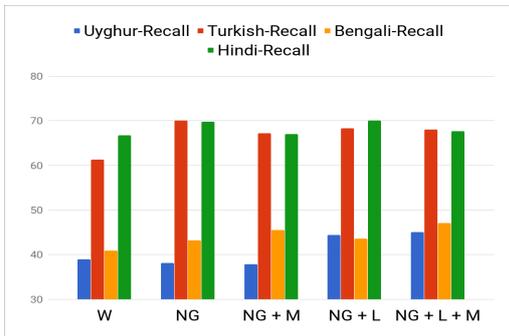

Figure 3: Recall for all languages (monolingual) **W**:Word, **NG**:Char-ngrams, **L**:Lemma, **M**:Morph

We find CT-Joint to be consistently better performing than CT-FineTune. Interestingly, the performance of CT-FineTune model converges to the monolingual performance. We hypothesize that the model forgets the pre-trained subword units as training progresses.

For CT-FineTune, the column *subword units* in Table 2 refers to the subword units which were pre-trained on a high resource related language. For example *lemma + morph* means lemma and morph embeddings are first pre-trained on the resouce-rich language and then used to initialize the respective lemma and morph representations for the low resource language.

**Monolingual Experiments:** Table 3 shows our results on all languages. We get +5.8 F1 points for Turkish, +4.8 F1 for Uyghur, +0.8 F1 for Hindi and +0.7 F1 for Bengali over the existing methods. We observe that a combination of character-ngrams, lemma and morphological properties gives the best performance for Uyghur and Bengali. Adding *morph* hurts in Turkish, in contrast to Hindi, where it helps. Section 5.3.3 discusses plausible reasons for this.

We report official NIST scores on the full evaluation set for Uyghur, as part of LORELEI Offical Retest. Additionally, we compare our results with the best results reported on the same LORELEI dataset. Results are seen in Table 4.

| Model | subword units | Uyghur | Bengali |
|---|---|---|---|
| CT-Joint | phoneme-ngrams + lemma + morph | 23.04 | 7.88 |
| | phoneme-ngrams + lemma | 23.24 | 7.62 |
| | phoneme-ngrams | 23.25 | 7.45 |
| CT-FineTune | lemma + morph | 23.71 | 7.58 |

Table 5: Transfer experiments for MT. Metric: BLEU. Uyghur transfer is from Turkish; Bengali transfer is from Hindi

| Model | subword units | Uyghur | Bengali |
|---|---|---|---|
| Ours | Char-ngrams + Lemma + Morph | 23.59 | **7.96** |
| | Char-ngrams + Lemma | **23.91** | 7.77 |
| | Char-ngrams + Morph | 23.27 | 7.88 |
| fastText | Char-ngrams | 23.24 | 7.91 |
| word2vec | Word | 23.31 | 6.64 |
| Random | No embedding | 23.51 | 6.23 |

Table 6: MT results for monolingual experiments. Metric: BLEU

### 5.3.3 Error analysis

We plot recall curves for all languages. As seen in Figure 3, adding subword units boosts the recall consistently across all languages, more so for Uyghur. For Turkish, *lemma* performs better than *lemma+morph*, perhaps because the morphological analyzer outputs so many redundant properties which reduce the distance between words that are not particularly similar. In contrast, *morph* helps and *lemma* hurts in Hindi, perhaps because the morph analyzer outputs only a small number of highly informative properties, but is a poor general-purpose lemmatizer.

We analyze our results for Uyghur language, as it was part of the LORELEI challenge and presents a situation close to a real-life application. We base our analysis on the unsequestered set since annotations for full test data are not released. There are 1,341 NE's in this set, 396 of which are covered by the word embeddings when trained with just monolingual corpus. One obvious advantage of jointly training with a resource-rich corpus is that coverage of NEs increases, as validated in our case where jointly training with Turkish corpus adds 114 more NEs.

Figure 4 shows ten named entities in two different embeddings (CT-Joint: *phoneme-

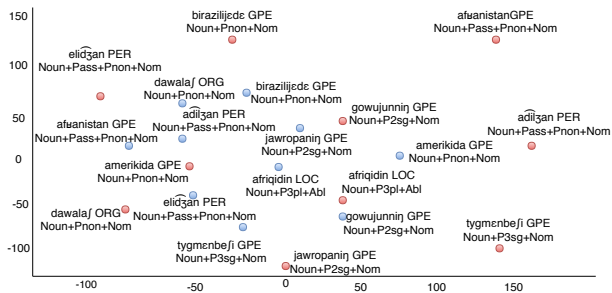

Figure 4: Two-dimensional PCA projection of select NEs from word embeddings for Uyghur—**CT-Joint** model trained with *phoneme-ngrams+lemma+morph* (blue) and monolingual model trained with *char-ngrams+lemma+morph*

ngrams+lemma+morph* and monolingual: *char-ngram+lemma+morph*). The difference is striking—in the monolingual condition, the NEs are widely dispersed, but in the bilingual condition, the NEs cluster together. This suggests that phonologically-mediated transfer through Turkish is resulting in embeddings in which NEs are close to one another, relative to monolingual Uyghur embeddings.

### 5.4 Machine Translation Task

In addition to NER, we test the performance of our proposed approaches on the MT task to test generality of our conclusions. We use XNMT toolkit

([Neubig et al., 2018](#)) to translate sentences from the low-resource language to English. We run similar transfer and monolingual experiments as done for NER. Due to space limitations, we use select subword combinations for the experiments, details of which can be found in Appendix. BLEU is used as the evaluation metric.

From Table 6, we observe that the combination of *character-ngrams* and *lemma* performs the best for Uyghur (+0.1) and the combination of *character-ngrams*, *lemma* and *morph* gives the best performance for Bengali (+1.7), over the *word* baseline, which demonstrates the importance of subword units for low-resource MT as well. One likely reason that the combination of character-ngrams and lemmas consistently show the best performance is that, together, they capture lexical similarity, which is more important to translation than the syntactic information captured by morphological inflection ("morph"). However, experiments using CT-Joint and CT-FineTune (Table 5) do not follow the same trend as that of NER. We hypothesize that this is because the MT models were trained on a training set that did not have translation pairs from the high resource language. As Qi et al. (2018) note, when training MT systems on a single language pair, it is less necessary for the embeddings to be coordinated across the languages.

## 6 Related Work

**Word Embedding Models:** Most algorithms for learning embeddings take inspiration from language modeling (Bengio et al., 2003), motivated by distributional hypothesis (Harris, 1954), and employ a shallow neural network to map the words into a low dimensional space. Pennington et al. (2014) built over the above local context window model by combining it with global matrix factorization (Levy and Goldberg, 2014). Recently, Peters et al. (2018) show significant gains across various tasks by learning word vectors as hidden states of a deep bi-directional language model. This was originally conceived for resource-abundant languages, hence it is as-of-yet unclear how generalizable they are to low-resource settings.

**Modeling subword information:** Various methods have validated the importance of modeling subword units in downstream tasks. Xu et al. (2016); Chen et al. (2015) experiment at the character level whereas Luong et al. (2013) use morphemes as a basic unit in recursive neural network (RNN) to get morphologically-aware word representations. Xu and Liu (2017) incorporate the morphemes' meanings as part of the word representation to implicitly model the morphological knowledge.

**Transfer learning:** Most recent works using transfer in low resource setting are coupled tightly with the downstream task. Jin and Kann (2017) use morpheme units for cross-lingual transfer in a paradigm completion task using sequence-to-sequence models. Tsai et al. (2016) employ a language-independent method for NER by grounding non-English phrases to English Wikipedia. Interestingly, Kim et al. (2017) use separate encoders for modeling language-specific and language-agnostic features for part-of-speech (POS) tagging, and make use of no cross-lingual resources.

## 7 Conclusion

In this paper, we explored two simple methods for cross-lingual transfer, both of which are task-independent and use transfer learning for leveraging subword information from resource-rich languages, especially through phonological and morphological representations. CT-Joint and CT-FineTune do not require morphological analyzers, but we have found that even a morphological analyzer built in 2-3 weeks can boost performance and is a worthwhile investment of resources. Preliminary evaluation on a separate task of MT reconfirms the utility of subword units and further research will reveal what these learned subword representations can contribute to other tasks.

## Acknowledgement

This work is sponsored by Defense Advanced Research Projects Agency Information Innovation Office (I2O). Program: Low Resource Languages for Emergent Incidents (LORELEI). Issued by DARPA/I2O under Contract No. HR0011-15-C0114. The views and conclusions contained in this document are those of the authors and should not be interpreted as representing the official policies, either expressed or implied, of the U.S. Government. The U.S. Government is authorized to reproduce and distribute reprints for Government purposes notwithstanding any copyright notation here on.

## Appendix

### MT Experimental Setup

We conduct the two sets of experiments, similar to the NER experiments:

1. Experiments where word embeddings are transferred from Uyghur and Bengali, using Turkish and Hindi as the respective high resource languages. Both the proposed models, CT-Joint and CT-FineTune, are used for experimentation with select subword units.

2. Monolingual experiments on the two low resource languages: Uyghur and Bengali, with select subword combinations.

### Data Processing

We use data, comprised of unlabeled corpora, training translation pairs (if provided), from the same sources as used for the NER experiments. The training data comprises of translation pairs between the source language and the target language, English. We create our own train-dev-test splits for the experiments as there are no official splits provided. The data splits can be seen in Table 7. The Uyghur corpus has 31 million tokens (extracted from a different set than the one used for NER) and the Turkish corpus about 40 million tokens. The Bengali corpus has 125 million tokens and we downsampled the original Hindi corpus to a comparable size.

| Lang. | Train | Dev | Test |
|---|---|---|---|
| Uyghur | 99379 | 994 | 994 |
| Bengali | 101523 | 1989 | 1988 |

Table 7: Sentences in train/dev/test set for MT.

### Model Setup

We train the model using 512-dimensional word embeddings, pre-trained using strategies described in the main text. We run each MT system 3 times and report the median score, as a way to control for variance in training.